\documentclass[runningheads]{llncs}
\usepackage{listings}
\usepackage[T1]{fontenc}
\usepackage{graphicx}
\usepackage{color}
\usepackage[title]{appendix}
\usepackage{multirow}
\usepackage{tabularx}
\usepackage{hyperref}

\begin{document}
\renewcommand\UrlFont{\color{blue}\rmfamily}
\urlstyle{rm}
\pagestyle{empty}

\title{Multi-Meta-RAG: Improving RAG for Multi-Hop Queries using Database Filtering with LLM-Extracted Metadata}
\titlerunning{Multi-Meta-RAG}
\author{Mykhailo Poliakov\orcidID{0009-0006-5263-762X} \and
Nadiya Shvai\orcidID{0000-0001-8194-6196}}
\authorrunning{M. Poliakov and N. Shvai}
\institute{National University of Kyiv-Mohyla Academy \\
\email{\{mykhailo.poliakov, n.shvay\}@ukma.edu.ua}}
\maketitle
\begin{abstract}
The retrieval-augmented generation (RAG) enables retrieval of relevant information from an external knowledge source and allows large language models (LLMs) to answer queries over previously unseen document collections. However, it was demonstrated that traditional RAG applications perform poorly in answering multi-hop questions, which require retrieving and reasoning over multiple elements of supporting evidence. We introduce a new method called Multi-Meta-RAG, which uses database filtering with LLM-extracted metadata to improve the RAG selection of the relevant documents from various sources applicable to the question. While database filtering is specific to a set of questions from a particular domain and format, we found that Multi-Meta-RAG greatly improves the results on the MultiHop-RAG benchmark. The code is available on \href{https://github.com/mxpoliakov/Multi-Meta-RAG}{GitHub}.

\end{abstract}

\section{Introduction} 

Large Language Models (LLMs) have shown remarkable language understanding and generation abilities \cite{llm:2022,llama:2023}. However, there are two main challenges: static knowledge \cite{augmented:2023} and generative hallucination \cite{hallucination:2023}. Retrieval-augmented generation \cite{rag:2020} is an established process for answering user questions over entire datasets. RAG also helps mitigate generative hallucination and provides LLM with new information on which it was not trained \cite{rag-reduce-hallucination:2021}. Real-world RAG pipelines often need to retrieve evidence from multiple documents simultaneously, a procedure known as multi-hop querying. Nevertheless, existing RAG applications face challenges in answering multi-hop queries, requiring retrieval and reasoning over numerous pieces of evidence \cite{multihoprag:2024}. In this paper, we present Multi-Meta-RAG: an improved RAG using a database filtering approach with LLM-extracted metadata that significantly improves the results on the MultiHop-RAG benchmark.
 
\section{Related works}

MultiHop-RAG \cite{multihoprag:2024} is a novel benchmarking dataset focused on multi-hop queries, including a knowledge base, questions, ground-truth responses, and supporting evidence. The news articles were selected from September 26, 2023, to December 26, 2023, extending beyond the knowledge cutoff of ChatGPT\footnote{gpt-3.5-turbo-0613} and GPT-4\footnote{gpt4-0613}. A trained language model extracted factual or opinion sentences from each news article. These factual sentences act as evidence for multi-hop queries. The selection method involves keeping articles with evidence that overlaps keywords with other articles, enabling the creation of multi-hop queries with answers drawn from numerous sources. Given the original evidence and its context, GPT-4 was used to rephrase the evidence, referred to as claims. Afterward, the bridge entity or topic is used to generate multi-hop queries. 

For example, \textit{"Did Engadget report a discount on the 13.6-inch MacBook Air before The Verge reported a discount on Samsung Galaxy Buds 2?"} is a typical query from the MultiHop-RAG dataset. Answering it requires evidence from \textit{Engadget} and \textit{The Verge} to formulate an answer. Also, it requires LLM to figure out the temporal ordering of events. MultiHop-RAG also has inference, comparison, and null (without correct answer) queries, in addition to the temporal query above. 

\begin{figure}[ht]
\includegraphics[width=\textwidth]{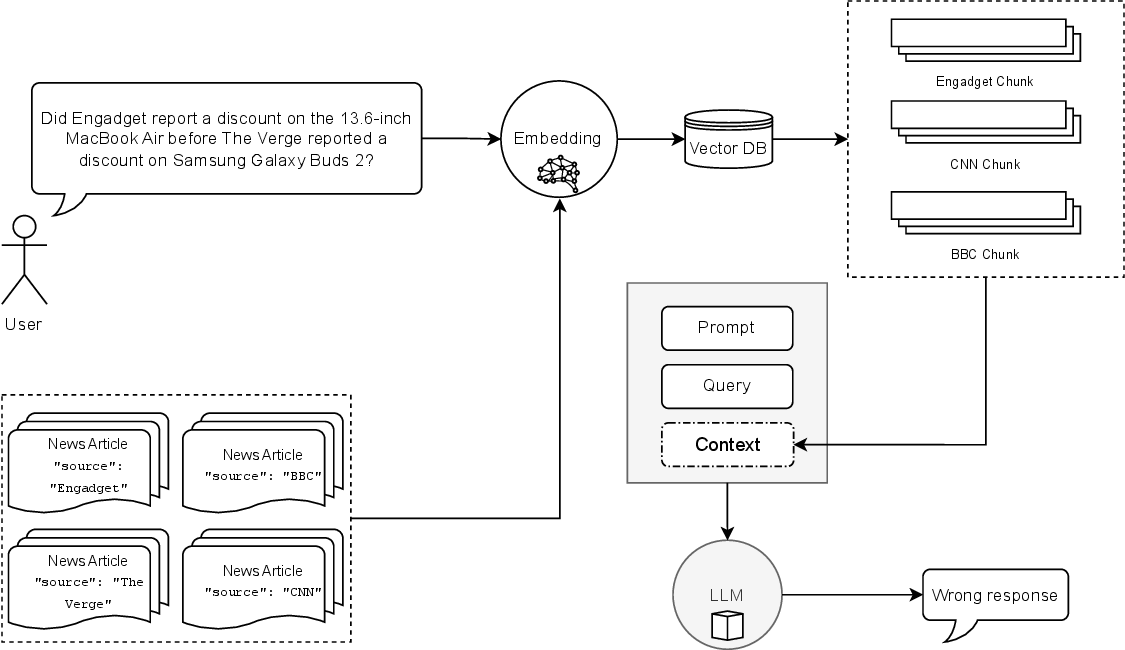}
\caption{A naive RAG implementation for MultiHop-RAG queries. RAG selects chunks from articles not asked in the example query, which leads to LLM giving a wrong response.} \label{rag-figure}
\end{figure}

\noindent In a typical RAG application, we use an external corpus that comprises multiple documents and serves as the knowledge base. Each document within this corpus is segmented into chunks. These chunks are then converted into vector representations using an embedding model and stored in a vector database. Given a user query, RAG typically retrieves the top-K chunks that best match the query. The retrieved chunks, combined with the query, are submitted to an LLM to generate a final response.

For the MultiHop-RAG benchmark, scraped articles act as a knowledge base for the RAG application tested. The problem is that a naive RAG application fails to recognize that the query asks for information from specific sources. Top-K chunks such as RAG retrieves often contain information from sources other than those mentioned in the query. Retrieved chunks might even miss relevant sources, leading to a wrong response, as depicted in Figure \ref{rag-figure}. 

Several popular benchmarks, such as HotpotQA \cite{hotpotqa:2018} and 2WikiMultiHopQA \cite{multi-qa-wiki:2020}, can be used for QA from multiple document sources. However, these datasets primarily focus on estimating LLM reasoning skills and do not highlight retrieving evidence from the knowledge base. Another problem is that they are based on Wikipedia, which means LLM's are already trained on the same data. 

Alternative solutions to tackle multi-hop queries include graph-based solutions like Graph RAG \cite{graphrag:2024}. While Graph RAG evaluates MultiHop-RAG dataset, it is used purely as a knowledge base for an independent question set. Another LLM assesses the responses for custom metrics such as comprehensiveness, diversity, empowerment, and directness instead of simple accuracy. 

\section{Multi-Meta-RAG}
\subsection{Extraction of Relevant Query Metadata with the LLM}
\begin{table}[ht]
\caption{Examples of extracted metadata filters using a few-shot prompt with corresponding queries. Correct usage of the \$nin operator for the last query can be noted.}\label{extracted-filters-table}
\begin{tabular}{p{0.35\textwidth} p{0.01\textwidth} p{0.61\textwidth}}
Query & & Extracted Filter \\ \hline \vspace{0.5pt}
Does the TechCrunch article report on new hiring at Starz, while the Engadget article discusses layoffs within the entire video game industry? & & 
\begin{lstlisting}[basicstyle=\scriptsize, aboveskip=5pt]
"source": {
    "$in": ["TechCrunch", "Engadget"]
}
\end{lstlisting} \\ \hline \vspace{0.5pt}
Did The Guardian's report on December 12, 2023, contradict the Sporting News report regarding the performance and future outlook of Manchester United? & &
\begin{lstlisting}[basicstyle=\scriptsize, aboveskip=0pt] 
"published_at": {
    "$in": ["December 12, 2023"]
},
"source": {
    "$in": ["The Guardian", "Sporting News"]
}
\end{lstlisting} \\ \hline \vspace{0.5pt}
Who is the individual facing a criminal trial on seven counts of fraud and conspiracy, previously likened to a financial icon but not by TechCrunch, and is accused by the prosecution of committing fraud for wealth, power, and influence? & &
\begin{lstlisting}[basicstyle=\scriptsize, aboveskip=20pt] 
"source": {
    "$nin": [
        "TechCrunch"
    ]
}
\end{lstlisting} \\ \hline
\end{tabular}
\end{table}

Each question in the MultiHop-RAG \cite{multihoprag:2024} benchmark follows a typical structure. Every query requests information from one or more sources of news. In addition, some temporal queries require news articles from a particular date. We can extract the query filter via helper LLM by constructing a few-shot prompt \cite{fewshot:2020} with examples of extracted article sources and publishing dates as a filter. The prompt template is provided in Appendix \ref{metadata-extraction-template}. We only run metadata extraction with ChatGPT\footnote{gpt-3.5-turbo-1106} because this additional RAG pipeline step must be quick and cheap. We found out that this step takes 0.7 seconds on average for one query.  

Two query metadata filter fields are extracted: article source and publication date. The complete filter is a dictionary with two fields combined. Samples of extracted metadata filters can be found in Table \ref{extracted-filters-table}. The primary filtering operator is \verb|$in|, the only operator provided in the examples in a few-shot prompt template. The LLM also correctly chooses a tiny fraction of the \verb|$nin| operator for some queries without an example. While LLM only used \verb|$in| and \verb|$nin| for article sources, the model sometimes chooses other operators like \verb|$lt| or \verb|$gt| for publication date for a fraction of temporal queries. Because the number of such queries is small, we decided to only use date filters with \verb|$in| and \verb|$nin| operators and a most frequent date format\footnote{strftime format code \%B \%-d, \%Y} for easier matching in the database. All queries have a source filter extracted, while the publishing date filter was extracted in 15.57\% of queries, while 22.81\% of queries of the MultiHop-RAG dataset are temporal.

\subsection{Improved Chunk Selection using Metadata Filtering}
\begin{figure}[ht]
\includegraphics[width=\textwidth]{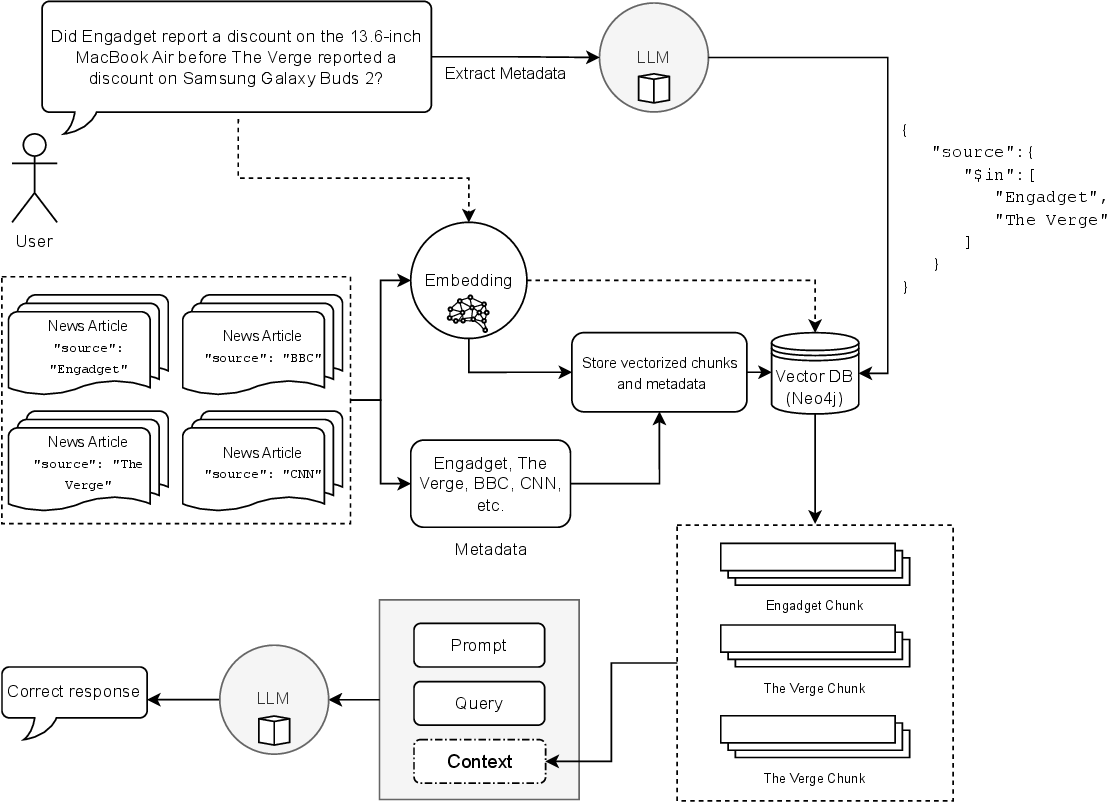}
\caption{Multi-Meta-RAG: an improved RAG with database filtering using metadata. Metadata is extracted via secondary LLM. With filtering, we can ensure top-K chunks are always from relevant sources with better chances of getting correct overall responses.} \label{improved-rag-figure}
\end{figure}
The extracted metadata could be used to enhance an RAG application (Figure \ref{improved-rag-figure}). We split the articles in the MultiHop-RAG \cite{multihoprag:2024} knowledge base into chunks, each containing 256 tokens using LLamaIndex \cite{llamaindex:2022} using a sentence splitter as in the original MultiHop-RAG implementation. We also picked a chunk overlap of 32, finding out that smaller chunk overlap leads to a better variety of unique chunks in the top-K selection than the original implementation, which used the LLamaIndex default of 200. We selected LangChain \cite{langchain:2022} Neo4j \cite{neo4j} vector store as a vector database as its index implementation recently\footnote{April 2024} started to support metadata filtering. We then convert the chunks using an embedding model and save the embeddings into a vector database with article metadata saved as node properties.

Likewise, in the retrieval stage, we transform a query using the same embedding model and retrieve the top-K most relevant chunks with the highest cosine similarity with the query embedding. We also filter the chunks with LLM-extracted metadata in the same stage. Similarly to MultiHop-RAG, we use a Reranker module (bge-reranker-large \cite{bge:2024}) to examine the retrieval performance. After retrieving 20 corresponding chunks using the embedding model and metadata filter, we select the top-K chunks using the Reranker.

\section{Results}
\subsection{Chunk Retrieval Experiment}
We selected two best-performing embedding models from the original MultiHop-RAG experiment for testing metadata filtering chunk retrieval performance, bge-large-en-v1.5 \cite{bge:2024}, and voyage-02 \cite{voyage}. The retrieved list of chunks is compared with the ground truth evidence associated with each query, excluding the null queries, as they lack corresponding evidence. For evaluation, we assume the Top-K chunks are retrieved and use metrics such as Mean Average Precision at K (MAP@K), Mean Reciprocal Rank at K (MRR@K), and Hit Rate at K (Hit@K). MAP@K measures the average precision of the top-K retrieval across all queries. MRR@K calculates the average reciprocal ranks of the first relevant chunk within the top-K retrieved set for each query. Hit@K measures the proportion of evidence that appears in the top-K retrieved set. The experiment (Table \ref{chunk-retrieval-table}) with RAG showed considerable improvement in both embeddings for all core metrics: MRR@10, MAP@10, Hits@10, and Hits@4. Most notably, for voyage-02, Hits@4 enhanced by 17.2\%. This improvement is important for practical  RAG systems, where the top-K retrieved should be as low as possible to account for context window limits and cost.
\begin{table}[ht]
    \caption{Chunk retrieval experiment results. Top-10 chunks are selected with bge-reranker-large after the top-20 chunks are found via the similarity search and database metadata filtering. A chunk size of 256 and a chunk overlap of 32 is used. We evaluate both Baseline RAG and Multi-Meta-RAG using an evaluation script provided in the MultiHop-RAG repository.}\label{chunk-retrieval-table}
    \def\arraystretch{1.5}
    \begin{tabularx}{\linewidth}{p{0.4\linewidth}XXXX}
        \hline
        \multirow{2}{*}{Embedding} & \multicolumn{4}{c}{Baseline RAG \cite{multihoprag:2024}} \\
        \cline{2-5}
        & MRR@10 & MAP@10 & Hits@10 & Hits@4 \\
        \hline
        bge-large-en-v1.5 (evaluated) & 0.6029 & 0.2687 & 0.7490 & 0.6661 \\
        voyage-02 (evaluated) & 0.6016 & 0.2619 & 0.7419 & 0.6630 \\
        \hline
    \end{tabularx}
    \begin{tabularx}{\linewidth}{p{0.4\linewidth}XXXX}
        \multirow{2}{*}{Embedding} & \multicolumn{4}{c}{Multi-Meta-RAG (ours)} \\
        \cline{2-5}
        & MRR@10 & MAP@10 & Hits@10 & Hits@4 \\
        \hline
        bge-large-en-v1.5 & 0.6574 & 0.3293 & 0.8909 & 0.7672 \\
        voyage-02 & \textbf{0.6748} & \textbf{0.3388} & \textbf{0.9042} & \textbf{0.792} \\
        \hline
    \end{tabularx}
\end{table}
\subsection{LLM Response Generation Experiment}
\begin{table}[ht]
    \caption{Overall generation accuracy of LLMs with MultiHop-RAG (top-6 chunks with voyage-02)}\label{models-table}
    \def\arraystretch{1.5}
    \begin{tabularx}{\linewidth}{p{0.25\linewidth}XXp{0.28\linewidth}}
        \hline 
        \multirow{2}{*}{LLM} & \multicolumn{3}{c}{Accuracy} \\
        \cline{2-4}
        & Ground-truth \cite{multihoprag:2024} & Baseline RAG \cite{multihoprag:2024} & Multi-Meta-RAG (ours) \\ 
        \hline
        {\scriptsize \textbf{GPT4} (gpt-4-0613)} & 0.89 & \textit{0.56} & \textbf{0.606} \\
         {\scriptsize \textbf{PaLM} (text-bison@001)} & 0.74 & \textit{0.47} & \textbf{0.608}\\
        \hline
    \end{tabularx}
\end{table}

As with embeddings, we picked two best-achieving LLMs on ground-truth chunks based on MultiHop-RAG initial experiments, GPT-4 and Google PaLM. We achieved substantial improvement in accuracy (Table \ref{models-table}) for both models compared to baseline RAG implementation. Google PaLM accuracy improved by 25.6\% from 0.47 to 0.608. GPT-4 results also show a 7.89\% increase from 0.56 to 0.63. The accuracy is calculated by checking if any word in an LLM-generated response is present in the correct gold answer for each question.

\begin{table}[ht]
    \caption{Generation accuracy of LLMs with MultiHop-RAG per question type (top-6 chunks with voyage-02)}\label{models-table-per-question-type}
    \def\arraystretch{1.5}
    \begin{tabularx}{\linewidth}{p{0.25\linewidth}XX}
        \hline 
        \multirow{2}{*}{Question Type} & \multicolumn{2}{c}{Accuracy} \\
        \cline{2-3}
        & \scriptsize \textbf{GPT4} (gpt-4-0613) & \scriptsize \textbf{PaLM} (text-bison@001) \\ 
        \hline
        {Inference} & 0.951 & 0.9203 \\
        {Comparison} & 0.382 & 0.5397 \\
        {Temporal} & 0.256 & 0.4545 \\
        {Null} & 0.9867 & 0.2492 \\
        \hline
    \end{tabularx}
\end{table}

Table \ref{models-table-per-question-type} shows the detailed evaluation results of different question types for GPT-4 and Google PaLM. Both models scored remarkable scores that exceeded 0.9 for inference queries. Google PaLM performs significantly better for comparison and temporal queries than GPT-4. However, PaLM struggles with Null questions, whereas GPT-4 achieves a near-perfect score. These results suggest that combining both models for different queries can be a valid strategy to increase overall accuracy further.

\section{Conclusion}
This paper introduces Multi-Meta-RAG, a method of improving RAG for multi-hop queries using database filtering with LLM-extracted metadata. Multi-Meta-RAG considerably improves results in chunk retrieval and LLM generation experiments while being relatively straightforward and explainable compared to alternative solutions, like Graph RAG \cite{graphrag:2024}.
\subsection{Limitations}
The proposed solution still has some limitations. Firstly, extracting metadata requires a set of queries from a particular domain and question format, as well as additional inference time. Secondly, it requires the manual creation of a prompt template that will extract the metadata from the query. Thirdly, while the improved results are encouraging, they still fall considerably below the results achieved by feeding LLM precise ground-truth facts.
\subsection{Future work}
Future work includes trying more generic prompt templates for metadata extraction using multi-hop datasets from different domains. In addition, testing alternative LLMs, like LLama 3.1 \cite{llama:2023}, on datasets with more recent cut-off dates is viable.

\subsubsection{\ackname} This research was partially funded by OpenAI Researcher Access Program (Application 0000005294).

\section*{Appendix}
\subsection*{Metadata Extraction Prompt Template} \label{metadata-extraction-template}
Given the question, extract the metadata to filter the database about article sources. Avoid stopwords.\\\rule{\textwidth}{0.1pt}
The sources can only be from the list: ['Yardbarker', 'The Guardian', 'Revyuh Media', 'The Independent - Sports', 'Wired', 'Sport Grill', 'Hacker News', 'Iot Business News', 'Insidesport', 'Sporting News', 'Seeking Alpha', 'The Age', 'CBSSports.com', 'The Sydney Morning Herald', 'FOX News - Health', 
'Science News For Students', 'Polygon', 'The Independent - Life and Style', 'FOX News - Entertainment', 'The Verge', 'Business Line', 'The New York Times', 'The Roar | Sports Writers Blog', 'Sportskeeda', 'BBC News - Entertainment \& Arts', 'Business World', 'BBC News - Technology', 'Essentially Sports', 'Mashable', 'Advanced Science News', 'TechCrunch', 'Financial Times', 'Music Business Worldwide', 'The Independent - Travel', 'FOX News - Lifestyle', 'TalkSport', 'Yahoo News', 'Scitechdaily | Science Space And Technology News 2017', 'Globes English | Israel Business Arena', 'Wide World Of Sports', 'Rivals', 'Fortune', 'Zee Business', 'Business Today | Latest Stock Market And Economy News India', 'Sky Sports', 'Cnbc | World Business News Leader', 'Eos: Earth And Space Science News', 'Live Science: The Most Interesting Articles', 'Engadget']\\\rule{\textwidth}{0.1pt}
Examples to follow:\\
Question: Who is the individual associated with the cryptocurrency industry facing a criminal trial on fraud and conspiracy charges, as reported by both The Verge and TechCrunch, and is accused by prosecutors of committing fraud for personal gain?\\
Answer: \{'source': \{'\$in': ['The Verge', 'TechCrunch']\}\}\\
Question: After the TechCrunch report on October 7, 2023, concerning Dave Clark's comments on Flexport, and the subsequent TechCrunch article on October 30, 2023, regarding Ryan Petersen's actions at Flexport, was there a change in the nature of the events reported?\\
Answer: \{'source': \{'\$in': ['TechCrunch']\}, 'published\_at': '\$in': \{['October 7, 2023', 'October 30, 2023']\}\}\\
Question: Which company, known for its dominance in the e-reader space and for offering exclusive invite-only deals during sales events, faced a stock decline due to an antitrust lawsuit reported by 'The Sydney Morning Herald' and discussed by sellers in a 'Cnbc | World Business News Leader' article?\\
Answer: \{'source': \{'\$in': ['The Sydney Morning Herald', 'Cnbc | World Business News Leader']\}\}\\\rule{\textwidth}{0.1pt}
If you detect multiple queries, return the answer for the first. Now it is your turn:\\
Question: <query>\\
Answer:
%
%
%
\bibliographystyle{splncs04}
\bibliography{bibliography}
\end{document}